%% file: main.tex
\documentclass[10pt,twocolumn,letterpaper]{article}
\usepackage[pagenumbers]{cvpr}
\input{preamble}

\title{OrTeN: Neural Inverse Graphics for Joint Terrain and Line-scan Camera Trajectory Reconstruction}
\hypersetup{pdftitle={OrTeN: Neural Inverse Graphics for Joint Terrain and Linescan Camera Trajectory Reconstruction}}
\input{authors}

\makeatletter
\g@addto@macro\@maketitle{\input{figures/baseline-overview/figure}}
\makeatother

\begin{document}
\maketitle
\input{sections/abstract}
\input{sections/introduction}
\input{sections/related_work}
\input{sections/method}
\input{sections/experiments}

\input{sections/conclusion}

{\small

\input{main.bbl}
}
\end{document}

%% file: preamble.tex
\usepackage{graphicx}
\usepackage{amsmath,amssymb}
\usepackage{booktabs}
\usepackage{xcolor}
\usepackage{tikz}
\usetikzlibrary{arrows.meta,positioning,calc}
\definecolor{cvprblue}{rgb}{0.21,0.49,0.74}
\usepackage[breaklinks,colorlinks,allcolors=cvprblue]{hyperref}
\usepackage[capitalize]{cleveref}
\newcommand{\R}{\mathbb{R}}

%% file: authors.tex
\author{
    Josef Biberstein\\Massachusetts Institute of Technology\\{\tt\small jxb@mit.edu}
\and Sertac Karaman\\Massachusetts Institute of Technology\\{\tt\small sertac@mit.edu}
}

%% file: figures/baseline-overview/figure.tex
\begin{minipage}{\textwidth}
\centering
\captionsetup{type=figure}
\includegraphics[width=\linewidth]{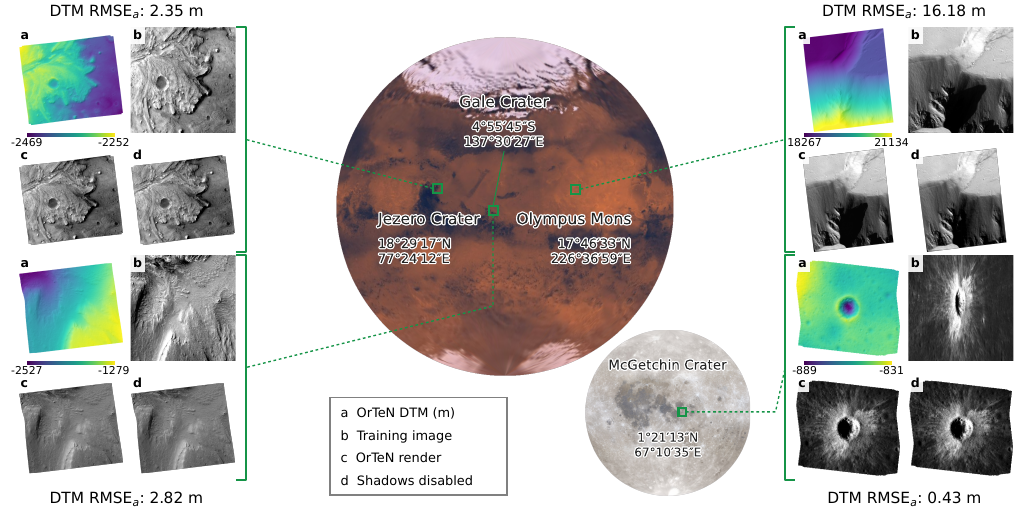}
\caption{\textbf{Summary of OrTeN DTM reconstructions.} 
OrTeN produces high-quality digital terrain maps (DTMs) from multi-view imagery without structural priors. 
Combining bundle adjustment, normalized cross-correlation, a semi-empirical photometric model, and geometric shadows allows OrTeN to reconstruct terrain reliably across multiple different instruments, planetary bodies, and lighting conditions.}
\label{fig:baseline-overview}
\end{minipage}
\par\vspace{1em}

%% file: sections/abstract.tex
\begin{abstract}
Digital terrain maps (DTMs) are used for landing site selection, terrain-relative navigation, and geological analysis in planetary exploration.
Existing state-of-practice reconstruction techniques for creating DTMs from satellite imagery use multi-view stereo, requiring complex pipelines for dealing with camera error, changing illumination, and images with limited disparity.
This work introduces Orbital Terrain Networks (OrTeN), a DTM construction method that uses neural inverse rendering to jointly optimize a textured height map and orbital camera corrections from line-scan satellite images.
OrTeN operates directly on calibrated line-scan images with pixel values expressed as radiance factor (I/F) and corrects both broad trajectory pose inconsistencies and high-frequency pointing jitter jointly with surface reconstruction during training.
Surface-projected patch normalized cross-correlation (NCC) enforces consistency between observations of the same terrain, refining an initial rough reconstruction.
A semi-empirical reflectance model and shadow rendering are used to fit surface appearance from multi-date imagery with differing solar illumination.
OrTeN requires only calibrated line-scan images and initial camera poses and does not use any structural priors including ranging/depth measurements or initial rough DTMs.
OrTeN is evaluated on four planetary imagery datasets.
Aligned terrain RMSEs are 2.35 m at Jezero Crater, 2.82 m at Gale Crater, 16.18 m at Olympus Mons, and 0.43 m at McGetchin Crater.
\end{abstract}

%% file: sections/introduction.tex
\section{Introduction}
\label{sec:intro}
Digital terrain maps (DTMs) are an important supporting data product in space exploration.
Accurate DTMs enable landing site selection and terrain-relative navigation (TRN) during entry, descent, and landing for modern autonomous space exploration missions such as Mars 2020~\cite{cheng2021reference,jung_digital_2020}.
They also provide the topographic context needed to interpret planetary landforms and their geological history~\cite{mcewen2007hirise}.

Current pipelines for orbital DTM production used in the space science community rely on classical multi-view stereo (MVS) techniques to infer terrain structure---features are detected and matched between radiometrically calibrated images, bundle adjustment (BA) is performed to refine camera poses, and stereo correspondences are triangulated to produce a 3D terrain model~\cite{kelvin_rodriguez_integrated_2024,beyer_ames_2018}.
MVS can produce high-quality maps suitable for mission operations, but a number of challenges inherent to both MVS itself and its application to satellite datasets complicate the process.
These challenges include the difficulty of getting reliable feature matches in the presence of shadows or low-texture regions, the line-scan geometry of most satellite images not admitting the epipolar constraints assumed by traditional MVS, and transients like dust storms or seasonal changes that alter the appearance of the terrain.
Correcting for these effects can require labor-intensive preprocessing and careful manual intervention, making MVS pipelines cumbersome to use.

Neural inverse graphics provides a methodology for scene reconstruction that can address many of the challenges faced by classical MVS pipelines.
For instance, satellite adaptations of neural radiance fields (NeRFs) model changing illumination and acquisition appearance while recovering scene geometry~\cite{derksen_shadow_2021,mari_sat-nerf_2022,mari_multi-date_2023}.
However, full 3D volumetric representations admit degenerate solutions---for instance, all scene density concentrated in front of cameras, explaining the observed images without recovering the true geometry---which can be particularly problematic when working with low-disparity orbital imagery where depth cues are weak.
While structural priors can help mitigate these degeneracies~\cite{zhang_sparsesat-nerf_2023,wagner_satgeo_2026}, these measurements are not always available or reliable.

OrTeN mitigates the shortcomings of classical MVS and volumetric neural reconstruction by optimizing a textured height map directly.
Critically, OrTeN requires no structural priors to achieve accurate terrain reconstruction from images and approximate camera poses.
It builds on Neural Terrain Maps (NTM)~\cite{biberstein2025ntm}, adding direct bundle adjustment of line-scan camera extrinsics jointly with training, a semi-empirical photometric model, and geometric shadow modeling.
The contributions of this work are:
\begin{enumerate}
    \item A pipeline for creating accurate DTMs through differentiable volume rendering of textured height maps including a first-order approximation of distance to the height map along the ray to improve performance at oblique angles.
    \item Joint terrain and line-scan camera optimization, including smooth trajectory adjustment and high-frequency jitter removal, and a semi-empirical, spatially varying photometric model including geometric shadow modeling with an image-space prior.
    \item An evaluation on four Mars and Moon scenes with varied relief and illumination, using the Context Camera (CTX) and Lunar Reconnaissance Orbiter Camera (LROC) Narrow Angle Camera (NAC).
\end{enumerate}
OrTeN achieves aligned height errors below the ground sampling distance (GSD) in three scenes and improves on the tested Ames Stereo Pipeline (ASP) reconstruction in those scenes.

%% file: sections/related_work.tex
\section{Related Work}
\label{sec:related}
\paragraph{DTM creation with MVS.}
A number of well-established pipelines exist for creating DTMs from orbital imagery using MVS, including the Ames Stereo Pipeline (ASP), satellite stereo pipeline (s2p), and COLMAP~\cite{beyer_ames_2018,de_franchis_automatic_2014,schonberger_structure--motion_2016,schonberger_pixelwise_2016}.
All of these methods use feature extraction and matching to constrain camera alignment before performing dense stereo matching to triangulate terrain.
Orbital imagery is often taken by linescan cameras, and therefore integration into standard MVS pipelines requires particular care because each detector line is acquired at a different pose.
This precludes the standard epipolar geometry assumptions used in perspective-camera MVS to make searching for correspondences straightforward.
Linescan epipolar loci are generally curved, so the global rectification used for ideal perspective cameras must be replaced by local or sensor-specific constructions~\cite{defranchis_on_2014,geng_generic_2020}.
ASP can also refine time-varying camera states to correct high-frequency jitter~\cite{asp_jitter}.
\par
Illumination changes, common in multi-date satellite data acquisitions, create an additional challenge for traditional MVS approaches.
Shadows and surface reflectance can change between acquisitions even when terrain is static and make feature selection and matching more difficult.
Therefore, masking shadows and other illumination-dependent artifacts is often necessary to improve the robustness of MVS matching.
Shadows can also provide important geometric information which can be exploited in reconstruction, and shape-from-shading methods exploit this to further refine existing DTMs~\cite{alexandrov_multiview_2018}.
OrTeN performs trajectory bundle adjustment, including jitter removal, jointly with training, and uses a semi-empirical photometric model with explicit geometric shadow handling to improve terrain reconstruction.

\paragraph{Neural reconstruction of orbital imagery.}
S-NeRF introduced illumination-aware neural rendering for satellite photogrammetry, and Sat-NeRF added rational polynomial coefficient (RPC) camera support and transient appearance modeling.
EO-NeRF built on these previous methods to establish a baseline framework for general-purpose scenes on Earth by incorporating geometrically consistent shadows with learned radiometric and camera corrections and transient appearance embeddings~\cite{derksen_shadow_2021,mari_sat-nerf_2022,mari_multi-date_2023}.
Season-NeRF separated seasonal appearance from Sun and viewing geometry~\cite{gableman_incorporating_2024}, while a semi-empirical bidirectional reflectance distribution function (BRDF) was incorporated into BRDF-NeRF~\cite{zhang_brdf_2025}.
Other methods address the geometry directly through depth supervision, surface constraints, or gravity-aligned and coarse-to-fine regularization~\cite{zhang_sparsesat-nerf_2023,qu_sat-mesh_2023,wagner_satgeo_2026}.
Gaussian splatting for satellite imagery has also emerged as an alternative to methods utilizing volume rendering, offering efficient, high-quality neural scene representations~\cite{aira_gaussian_2025,bournez_eogs_2025}.
\par
More recent works have begun to focus on large-scale terrain reconstruction using neural inverse graphics.
Both NTM and LNEM introduce direct rendering of linescan imagery for terrain reconstruction~\cite{biberstein2025ntm,lee_lnem_nodate}.
For descent imagery, de Almeida \etal\ incorporate a neural height field into reconstruction from wide-angle cameras~\cite{almeida_neural_2026}.
OrTeN eschews structural priors and retains the explicitly parameterized terrain representation, as is done in NTM, and adds ray-distance-based opacity and improved regularization with normalized cross-correlation (NCC).
Its coarse-to-fine optimization follows the principle established by BARF: low-frequency structure should stabilize camera alignment before fine appearance is exposed~\cite{lin2021barf}.

%% file: sections/method.tex
\section{Methodology}
\label{sec:method}
\input{figures/hero}
OrTeN represents the scene with a height field $h_\theta:\R^2\rightarrow\R$ in a local east--north--up (ENU) frame and learned surface appearance (\cref{fig:hero}).
The surface is $\mathbf{x}=(x,y,h_\theta(x,y))$, and unit-direction camera rays are given by $\mathbf r(t)=\mathbf o+t\mathbf d$.
The origin is constructed as the common look point of all cameras, the horizontal bounds are set by the image footprints on the horizontal plane, enlarged by a factor of 1.5, and the vertical bounds are set by global planetary height limits.

\subsection{Rendering a height field}
\label{sec:opacity}
OrTeN follows the lineage of NeuS, which derives opacity from the decrease of a logistic cumulative distribution function (CDF) along a ray through an implicit surface~\cite{wang_neus_2023}.
Applying this construction to distance along the ray rather than vertical height residual gives
\begin{align}
 \alpha_j&=\left[1-\frac{\Phi_\beta(q(t_{j+1}))}{\Phi_\beta(q(t_j))}\right]_+,\label{eq:alpha}\\
 T_j&=\prod_{k<j}(1-\alpha_k),\qquad w_j=T_j\alpha_j.\label{eq:weights}
\end{align}
Here $\Phi_\beta(q)=(1+e^{-q/\beta})^{-1}$, $[z]_+=\max(z,0)$, $\alpha_j$ is interval opacity, $T_j$ is transmittance to that interval, and $w_j$ is its rendering weight.
The CDF ratio is evaluated on a log scale to improve numerical stability.
Sampling locations are chosen with a proposal network as is done in Nerfacto~\cite{tancik_nerfstudio_2023}.

To obtain $q$, the effect of terrain slope and viewing direction on the height residual is considered:
\begin{align}
 f(t)&=r_z(t)-h_\theta(\mathbf r_{xy}(t)),\\
 f'(t)&=d_z-\nabla h_\theta(\mathbf r_{xy}(t))^\top\mathbf d_{xy},\\
 q(t)&=\frac{f(t)}{\max(-f'(t),10^{-3})}.\label{eq:q1}
\end{align}
Subscripts $xy$ and $z$ select horizontal and vertical components.
For a plane, $q(t)=t_\star-t$ when the denominator clamp is inactive, where $t_\star$ is the intersection distance.
This formulation ensures the learned width $\beta>0$ has the same metric meaning across terrain slopes and viewing angles.
Height derivatives are calculated by finite differences using a first-order formulation.

\subsection{Calibrated I/F image formation}
\label{sec:formation}
Training uses calibrated scalar radiance factor $I/F=\pi L/F_\odot$ images, where $L$ is observed radiance and $F_\odot$ is solar irradiance at the body~\cite{bell_iii_calibration_2013,robinson2010lroc}.
Albedo $a_\psi$ is aggregated over the ray samples:
\begin{equation}
 A=\sum_jw_j,\qquad
 \bar a=\frac{\sum_jw_j a_\psi(\mathbf r_{xy}(\bar t_j))}{\max(A,10^{-8})},
 \label{eq:appearance}
\end{equation}
where $\bar t_j$ is the interval midpoint.
A second network head predicts the spatial Lunar--Lambert mixture $m_\phi(x,y)$, integrated in the same way to give $\bar m$.
The normalized, weighted terrain normal $\mathbf n$ is used to compute positive incidence and emission cosines $\mu_0=[\mathbf n^\top\mathbf s]_+$ and $\mu=[\mathbf n^\top(-\mathbf d)]_+$, with $\mathbf s$ pointing toward the Sun.
The direct-light factor is
\begin{align}
 F(\mu_0,\mu,m)&=(1-m)\mu_0+\frac{2m\mu_0}{\max(\mu_0+\mu,10^{-6})},\\
 D&=\frac{F(\mu_0,\mu,\bar m)}{F(\cos30^\circ,1,\bar m)}.
\end{align}
This interpolates between Lambert and Lommel--Seeliger responses~\cite{mcewen_photometric_1991}.

\paragraph{Geometric shadows.}
Starting from the weighted camera-ray depth, a surface intersection is refined and traced toward the Sun until the ray leaves the scene bounds.
Solar interval opacities $\alpha_k^\odot$ are converted to visibility using
\begin{equation}
 V=V_{\rm att}\prod_k(1-\alpha_k^\odot),\qquad
 V_\eta=(1-\eta)V_{\rm att}+\eta V.
 \label{eq:visibility}
\end{equation}
Here $V_{\rm att}$ smoothly suppresses light on back-facing slopes and is zero when the Sun is below the local horizon.
The startup factor $\eta$ ramps cast shadows in over 500 updates.
Terrain outside the bounds cannot cast shadows in this model.

OrTeN uses a semi-empirical image formation model inspired by EO-NeRF's shadow-fill and affine brightness approach~\cite{mari_multi-date_2023}.
The predicted image for acquisition $i$ is
\begin{equation}
 \widehat y_i=g_i\bigl[V_\eta D+(1-V_\eta)a_{\rm amb}(\mathbf s)\bigr]\bar a+b_i.
 \label{eq:iof}
\end{equation}
The Sun-dependent ambient factor $a_{\rm amb}\in(0,1)$ fills shadows without inheriting the direct-light angular response.
Scalar gains $g_i=e^{u_i-\bar u}$ and offsets absorb acquisition-wide brightness differences, with $\bar u=M^{-1}\sum_i u_i$ fixing the geometric mean gain to one across $M$ acquisitions.
The same image model is used throughout training.

Histogram-valley shadow detection~\cite{adeline_shadow_2013} is adopted to provide an initial image-space shadow estimate at startup.
Only sufficiently dark, distinct pixels are accepted, and mask erosion is applied to remove boundary transients.
For an ordinary rendering batch $\mathcal B$, a regularizer is applied:
\begin{equation}
 \mathcal L_{\rm shadow}=|\mathcal B|^{-1}\sum_{p\in\mathcal B}z_p V_p^2,
 \label{eq:shadow-prior}
\end{equation}
where $z_p=1$ marks a detected core and $z_p=0$ means unknown.
This loss is one-sided, so unlabeled pixels are not treated as either shadow or sunlit.

\subsection{Linescan bundle adjustment}
\label{sec:ba}
A linescan pixel $(\ell,s)$ has nominal origin $\mathbf o_i^0(\ell)$ and direction $\mathbf d_i^0(\ell,s)$.
Intrinsics and distortion are retained, and corrections are optimized in the local scene frame:
\begin{align}
 \mathbf o_i&=\mathbf o_i^0+\Delta\mathbf p_i(u),\\
 \mathbf d_i&=\exp([\Delta\boldsymbol\omega_i(u)]_\times)\mathbf d_i^0,\\
 \Delta\mathbf p_i(u)&=s_p\sum_{k=0}^3B_k(u)\mathbf P_{ik},\\
 \Delta\boldsymbol\omega_i(u)&=s_\omega\left[\sum_{k=0}^3B_k(u)\mathbf W_{ik}+r\mathbf J_i(u)\right].
 \label{eq:camera}
\end{align}
Here $u\in[0,1]$ is normalized acquisition time and $B_k$ are cubic Bernstein functions.
The matrix $[\boldsymbol\omega]_\times$ implements the cross product with $\boldsymbol\omega$, and its exponential gives the corresponding rotation.
The learned controls $\mathbf P_{ik},\mathbf W_{ik}$ use physical scales $s_p,s_\omega$ which are constrained by soft priors based on expected trajectory error~\cite{cheng2021reference}.
During the refinement phase (\cref{sec:architecture}), the ramp $r$ introduces angular jitter $\mathbf J_i$, represented by a cubic B-spline with 64 intervals per acquisition.
Its global cubic component is removed so it cannot duplicate the smooth trajectory correction.

\subsection{Surface-guided patch consistency}
\label{sec:ncc}
In addition to uniformly sampled rays, at each training iteration $5\times5$ patches are sampled in source images and projected onto the current height field through the refined cameras.
Each patch point is then projected from the surface into other acquisitions, where bilinear interpolation is used to calculate observed intensities.
For accepted source--neighbor pairs, the normalized cross-correlation loss is
\begin{equation}
 \mathcal L_{\rm NCC}=\mathbb E_{P,i,j}\left[
 1-\frac{\langle P_i-\bar P_i,P_j-\bar P_j\rangle}
 {\|P_i-\bar P_i\|_2\,\|P_j-\bar P_j\|_2}\right].
 \label{eq:ncc}
\end{equation}
Here $P_i,P_j$ contain the observed patch intensities and $\bar P_i,\bar P_j$ are their means.
$\langle \cdot,\cdot\rangle$ is the inner product.
Invalid, occluded, low-contrast, and nearly parallel view pairs are rejected, and up to three of the best eligible neighbors are retained per source patch.
NCC is active in both roughing and refinement (\cref{sec:architecture}).

\subsection{Architecture and losses}
\label{sec:architecture}
\input{figures/architecture/figure}
\cref{fig:architecture} shows the overall network architecture.
Training progresses from images downsampled by $4\times$ to $2\times$, then to native resolution.
Active encoding levels are adjusted to approximately match the current image resolution.
There are two training stages: roughing and refinement.
Roughing ends when terrain and camera motion stabilize or the roughing update limit is reached.
Refinement lowers the geometry learning rate, increases NCC sampling, and introduces jitter and local regularization.

Let $s_y=\sqrt{\mathbb E_{\rm train}[y^2]}$ be the RMS brightness of all training pixels.
The rendering loss is
\begin{equation}
 \mathcal L_{I/F}=\mathbb E_{p\in\mathcal B}\left[(\widehat y_p-y_p)^2/s_y^2\right].
\end{equation}
Albedo starts exactly at $s_y$, ambient at 0.2, and gains and offsets at one and zero.
The remaining surface and appearance penalties are
\begin{align}
 \mathcal L_{\rm bend}&=\mathbb E_{\mathbf x}\left[s_h^2\|\mathbf H_{s_h}h_\theta(\mathbf x)\|_F^2\right],\\
 \mathcal L_{\rm width}&=\log\bigl(1+(\beta/g_0)^2\bigr),\\
 \mathcal L_{\rm radio}&=\mathbb E_i\left[((u_i-\bar u)/0.1)^2+(b_i/0.01)^2\right],\\
 \mathcal L_{\rm amb}&=\mathbb E_{\mathbf s}\left[\bigl[\operatorname{logit}a_{\rm amb}(\mathbf s)-\log9\bigr]_+^2\right].
\end{align}
Here $\mathbf H_{s_h}$ is the finite-difference Hessian.
In order, these losses regularize surface bending, control the rendering width, enforce radiometric calibration, and constrain the ambient illumination.
The fixed scale $g_0$ is the initial training-image GSD.
The optional ambient penalty acts only when the ambient factor exceeds 0.9.

Camera regularization penalizes the dimensionless trajectory controls and, in refinement, the sampled jitter motion.
Let $\mathbf c_i(u)$ concatenate the trajectory position and rotation corrections divided by $s_p$ and $s_\omega$.
Its second derivative penalizes trajectory curvature:
\begin{align}
 \mathcal L_{\rm camera}={}&\lambda_C\mathbb E_{i,k}(\|\mathbf P_{ik}\|_2^2+\|\mathbf W_{ik}\|_2^2)\nonumber\\
 &+\lambda_C\mathbf1_{r>0}\mathbb E_{i,u}\|\mathbf J_i(u)\|_2^2\nonumber\\
 &+\lambda_K\mathbb E_i\int_0^1\|\mathbf c_i''(u)\|_2^2du.
 \label{eq:camera-prior}
\end{align}
For each camera or solar ray, the proposal loss is the histogram-envelope objective used by Nerfstudio~\cite{tancik_nerfstudio_2023}:
\begin{equation}
 \mathcal L_{\rm env}=\sum_l\mathbb E_j
 \frac{[w_j-W_{lj}]_+^2}{w_j+10^{-7}}.
\end{equation}
Here $W_{lj}$ is proposal level $l$'s total mass in bins overlapping final interval $j$.
The camera and solar proposal losses are summed as $\mathcal L_{\rm proposal}$.
The complete objective is
\begin{align}
 \mathcal L={}&\mathcal L_{I/F}+\eta\lambda_N\mathcal L_{\rm NCC}
 +\lambda_W\mathcal L_{\rm width}+\lambda_R\mathcal L_{\rm radio}\nonumber\\
 &+\mathcal L_{\rm camera}+\mathcal L_{\rm proposal}\nonumber\\
 &+r\left(\lambda_B\mathcal L_{\rm bend}
 +\lambda_S\mathcal L_{\rm shadow}+\lambda_A\mathcal L_{\rm amb}\right).
 \label{eq:loss}
\end{align}
The refinement ramp $r$ is zero during roughing and rises over 500 updates after transition.

%% file: figures/hero.tex
\begin{figure*}[t]
\centering
\includegraphics[width=\textwidth]{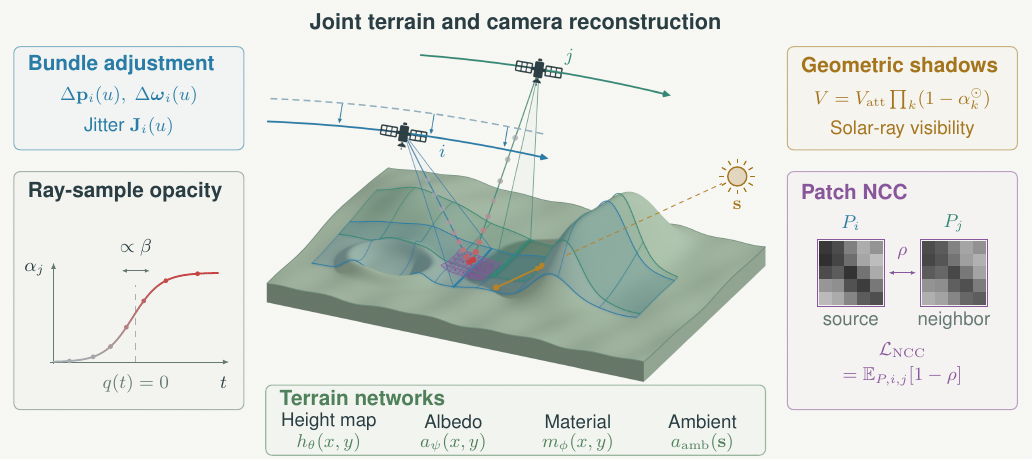}
\caption{
\textbf{Key elements of the OrTeN pipeline.}
Two linescan acquisitions $i,j$ observe overlapping terrain swaths.
The dashed trajectory indicates the original extrinsics, which are corrected during bundle adjustment.
Bundle adjustment refines position and attitude, including angular jitter.
Terrain networks predict height, albedo, material mixture, and a Sun-dependent ambient factor.
Gray-to-red camera-ray samples illustrate interval opacity $\alpha_j$, with transition width controlled by $\beta$.
Sun-directed opacity integration determines visibility $V$.
Source image patches are lifted onto the current height field and compared with neighbor observations through normalized cross-correlation $\rho$, constraining terrain and cameras jointly.
}
\label{fig:hero}
\end{figure*}

%% file: figures/architecture/figure.tex
\begin{figure}[t]
\centering
\resizebox{\linewidth}{!}{\begin{tikzpicture}[x=1cm,y=1.1cm,
 font=\sffamily\fontsize{8}{9.5}\selectfont,
 >={Latex[length=1.5mm,width=1.2mm]}, line width=.55pt,
 flow/.style={->,draw=cvprblue!85!black,line width=.85pt},
 relu/.style={->,draw=orange!85!black,line width=1.05pt},
 sigmoid/.style={->,draw=magenta!75!black,line width=1.05pt},
 learned/.style={draw=cvprblue!75!black,fill=cvprblue!8,rounded corners=2pt,align=center,inner sep=4pt},
 parameter/.style={draw=red!50!black,fill=red!6,rounded corners=2pt,align=center,inner sep=4pt},
 operation/.style={draw=black!45,fill=black!3,rounded corners=2pt,align=center,inner sep=4pt},
 branch/.style={text width=3.22cm},
 layer/.style={draw=cvprblue!75!black,fill=cvprblue!18,minimum width=.36cm,minimum height=.44cm,inner sep=0pt,font=\sffamily\fontsize{7}{8}\selectfont}]
\node[parameter,branch,minimum height=.75cm] (camera) at (-2,0) {\textbf{Camera controls}\\Trajectory + spline jitter};
\node[operation,branch,minimum height=.75cm] (ray) at (2,0) {\textbf{Linescan camera rays}\\$\mathbf r(t)=\mathbf o_i+t\mathbf d_i$};
\draw[flow] (camera.east)--(ray.west);
\node[align=center,inner sep=1pt] (xy) at (0,-.95) {Horizontal query $(x,y)=\mathbf r_{xy}(t)$};
\draw[flow] (ray.south)|-($(xy.north)+(0,.20)$)--(xy.north);
\node[learned,branch] (encoding) at (0,-1.75) {\textbf{Shared XY encoding}\\2D hash: $16\times2$ features};
\draw[flow] (xy.south)--(encoding.north);
\node[learned,text width=7.22cm,minimum height=1.05cm] (trunk) at (0,-2.95) {};
\node at (0,-2.62) {\textbf{Shared geometry and appearance trunk}};
\foreach \x/\w/\id in {-1.35/64/t1,-.45/64/t2,.45/64/t3,1.35/65/t4}
  \node[layer] (\id) at (\x,-3.10) {\w};
\foreach \a/\b in {t1/t2,t2/t3,t3/t4}
  \draw[relu] (\a.east)--(\b.west);
\draw[flow] (encoding.south)--(trunk.north);
\coordinate (split) at (1.35,-3.85);
\draw[draw=cvprblue!85!black,line width=.85pt] (t4.south)--(split);
\node[operation,branch,minimum height=.85cm] (height) at (-2,-4.55) {\textbf{Metric height mapping}\\1 output $\mapsto h_\theta(x,y)$};
\draw[sigmoid] (split)-| node[pos=.4,above] {} (height.north);
\node[learned,branch,minimum height=1.70cm] (heads) at (2,-4.80) {};
\node at (2,-4.16) {\textbf{Appearance heads}};
\node[anchor=west] at (.40,-4.63) {Albedo};
\node[anchor=west] at (.40,-5.22) {Material};
\foreach \x/\y/\w/\id in {1.75/-4.63/64/a1,2.50/-4.63/1/a2,1.75/-5.22/32/m1,2.50/-5.22/1/m2}
  \node[layer] (\id) at (\x,\y) {\w};
\draw[relu] (a1.east)--(a2.west);
\draw[relu] (m1.east)--(m2.west);
\node (a) at (3.22,-4.63) {$a_\psi$};
\node (m) at (3.30,-5.22) {$m_\phi$};
\draw[sigmoid] (a2.east)--(a.west);
\draw[sigmoid] (m2.east)--(m.west);
\draw[flow] (split)-| node[pos=.4,above] {} (heads.north);
\node[operation,branch,minimum height=1.05cm] (weights) at (-2,-6.45) {\textbf{Ray-distance opacity}\\$q_1\mapsto\alpha_j\mapsto w_j$\\from $h_\theta,\nabla h_\theta,\mathbf r,\beta$};
\node[operation,branch,minimum height=1.05cm] (appearance) at (2,-6.45) {\textbf{Integrated appearance}\\$\{w_j,a_j,m_j\}\mapsto A,\bar a,\bar m$\\Normalized by $A$};
\draw[flow] (height.south)--(weights.north);
\draw[flow] (heads.south)--(appearance.north);
\draw[flow] (weights.east)--(appearance.west);
\node[parameter,branch,minimum height=.65cm] (width) at (-2,-7.65) {\textbf{Rendering width}\\Bounded scalar $\beta>0$};
\node[parameter,branch,minimum height=.65cm] (radio) at (2,-7.65) {\textbf{Acquisition calibration}\\$g_i=\exp(u_i-\bar u),\quad b_i$};
\draw[flow] (width.north)--(weights.south);
\node[learned,text width=7.22cm,minimum height=1.20cm] (ambient) at (0,-8.95) {};
\node at (0,-8.58) {\textbf{Sun-only ambient network}};
\node (sun) at (-2.45,-9.12) {$\mathbf s\mapsto\mathrm{SH}_4(\mathbf s)$};
\foreach \x/\w/\id in {-.85/32/s1,0/32/s2,.85/1/s3}
  \node[layer] (\id) at (\x,-9.12) {\w};
\draw[flow] (sun.east)--(s1.west);
\draw[relu] (s1.east)--(s2.west);
\draw[relu] (s2.east)--(s3.west);
\node (amb) at (2.25,-9.12) {$a_{\rm amb}(\mathbf s)$};
\draw[sigmoid] (s3.east)--(amb.west);
\node[operation,text width=7.22cm,minimum height=1.20cm] (render) at (0,-10.70) {\textbf{Geometric shading and calibrated image formation}\\$h_\theta,\mathbf s\mapsto V_\eta,\mathbf n,\quad \bar m,\mathbf n,\mathbf s,\mathbf d\mapsto D$\\$\widehat y_i=g_i[V_\eta D+(1-V_\eta)a_{\rm amb}]\bar a+b_i$};
\draw[flow] (weights.west)--(-4.02,-6.45)|-(render.west);
\draw[flow] (appearance.east)--(4.02,-6.45)|-(render.east);
\draw[flow] (radio.east)--(4.02,-7.65);
\draw[flow] (ambient.south)--(render.north);
\draw[relu] (-3.85,-11.70)--(-3.30,-11.70);
\node[anchor=west] at (-3.23,-11.70) {ReLU};
\draw[sigmoid] (-1.45,-11.70)--(-.90,-11.70);
\node[anchor=west] at (-.83,-11.70) {Sigmoid};
\draw[flow] (1.40,-11.70)--(1.95,-11.70);
\node[anchor=west] at (2.02,-11.70) {No activation};
\end{tikzpicture}}
\caption{\textbf{OrTeN's architecture.} 
Learned networks are blue, optimized parameters are red, and analytic operations are gray.
Layer boxes give output dimensions, and arrow colors indicate activations.
One spatial encoding~\cite{muller_instant_2022} feeds the height, albedo, and Lunar--Lambert material outputs.
The ambient network uses a fourth-order spherical harmonic (SH) encoding of the Sun direction.
Height also supplies normals and solar visibility $V_\eta$.
Proposal sampling is omitted for clarity.}
\label{fig:architecture}
\end{figure}
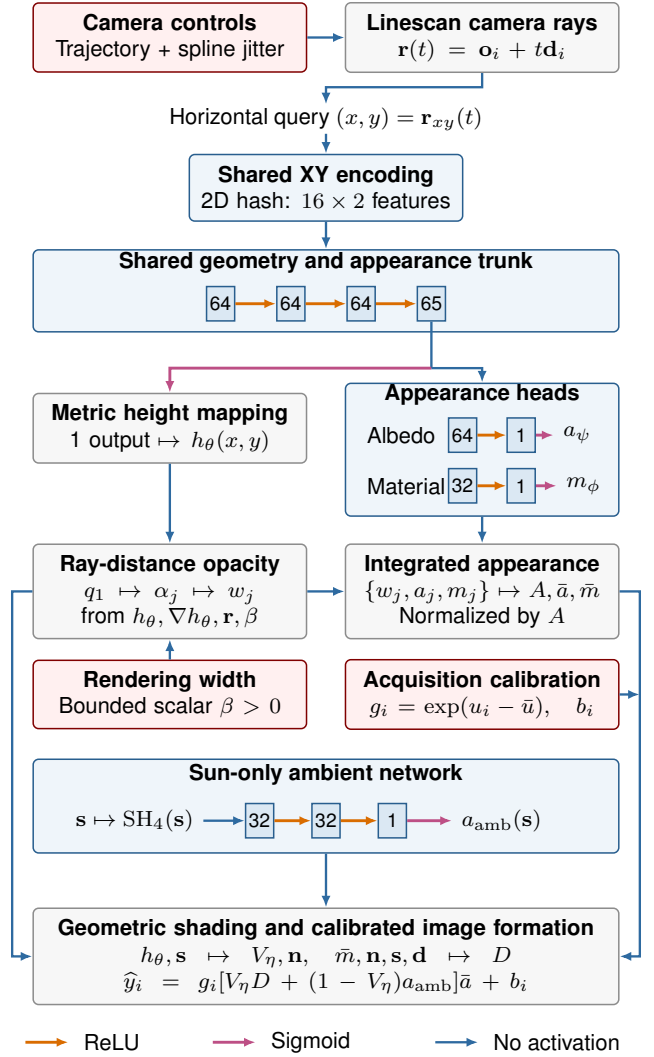

%% file: sections/experiments.tex
\section{Experiments and Results}
\label{sec:experiments}

\subsection{Datasets}
Four lunar and Martian datasets were selected for their varied terrain and illumination and the availability of reference DTMs.
Jezero Crater, Gale Crater, and Olympus Mons each contain fifteen $1024\times1024$ CTX crops.
McGetchin Crater contains ten LROC NAC crops of the same size.
Images are calibrated to I/F using the Integrated Software for Imagers and Spectrometers (ISIS) and paired with initial line-scan camera states.
\Cref{tab:datasets} gives additional details on the instruments and reference DTM quality for each dataset.
The Olympus dataset reference uses a DTM separately reconstructed from the same image collection because a high-quality engineering DTM was unavailable.
\input{tables/datasets}

\subsection{Evaluation protocol and metrics}
\label{sec:metrics}
Metrics for each scene are calculated over a fixed reference footprint that is covered by at least five initial cameras.
To resolve the gauge freedom inherent in the bundle adjustment process, reconstructed DTMs are aligned to the reference DTM using a six-degree-of-freedom rigid transformation.
No scale change, local warping, or separate alignment for different metrics is applied.
For aligned height errors $e_p=\widehat h_a(p)-h_{\rm ref}(p)$ on $N$ valid posts, the metrics are
\begin{equation}
    \mathrm{RMSE}_a=\sqrt{N^{-1}\sum_p e_p^2},\qquad
    \mathrm{MAE}_a=N^{-1}\sum_p |e_p|.
\end{equation}
The subscript $a$ denotes metrics applied to rigidly aligned DTMs, so these scores measure relative shape rather than absolute geolocation.

Slope is evaluated using a local quadratic surface fit calculated over a $9\times9$ sliding window.
This follows established practices in terrain analysis~\cite{gao2012slope} and reduces sensitivity to reference DTM artifacts.
Slope RMSE$_a$ is the RMS difference between slope angles in degrees, under the same rigid alignment.
Although OrTeN permits direct network differentiation, the reported comparison differentiates polynomial fits to every method's raster export to match spatial scale.
Height RMSE and MAE use unsmoothed elevations.

One $64\times64$ block per acquisition, at the full image resolution, is held out during training for perceptural metric calculation.
This is necessary because of learned image-dependent radiometric calibration.
PSNR is $10\log_{10}(R^2/\mathrm{MSE})$ where $R$ is the 99th minus 1st percentile of the pooled held-out observations per scene. 
SSIM uses the same range and $11\times11$ Gaussian windows ($\sigma=1.5$ pixels), averaged over valid window centers.

\subsection{Training details}
All training runs use a single seed, are limited to 15,000 iterations, and use a batch size of 8,192 rays.
The same configuration for OrTeN is used across all four scenes.
This includes enabling trajectory BA, jitter adjustment, and the detected-shadow prior with weight 0.01.
The roughing stage iteration cap is set to 8,000 updates.

Comparisons with depth-supervised EO-NeRF~\cite{mari_multi-date_2023} and ASP~\cite{beyer_ames_2018} are reported.
EO-NeRF uses the same camera-derived scene bounds as OrTeN, 8,192 rays per update, 128 depth samples per ray, and fixed cameras.
Its depth targets come from the unaligned ASP stereo mosaics, without reference-terrain supervision or registration.
ASP uses image-based BA, dense stereo, and jitter refinement anchored to its own stereo terrain estimates.
The same ASP pipeline is used for all scenes.

Mean setup-plus-training times are 52.7/41.2 minutes for OrTeN on NVIDIA RTX 6000 Ada Generation/RTX PRO 6000 Blackwell GPUs, respectively.
These include periodic evaluation and checkpointing, exclude final evaluation, and average different scene/configuration mixes across GPUs.
OrTeN uses Adam with a 500-step warmup to learning rates of $3\times10^{-3}$ for geometry and appearance, $10^{-3}$ for cameras, and $10^{-2}$ for proposal networks.
Geometry decays during roughing, then starts refinement at most $1.5\times10^{-4}$ and decays to $5\times10^{-5}$.
Appearance and camera rates finish at $3\times10^{-4}$ and $10^{-5}$.

\begin{figure}[!t]
\centering
\includegraphics[width=\columnwidth]{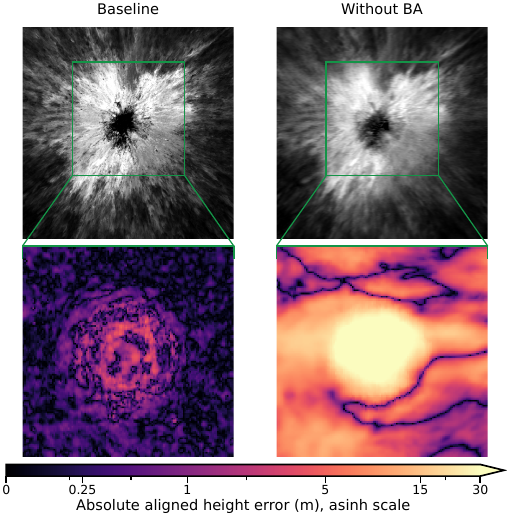}
\caption{\textbf{McGetchin Crater with and without BA.} 
Renderings and crater errors use shared scales. 
Boxes use the baseline camera. 
RMSE$_a$: 0.43 / 10.36 m. 
PSNR: 35.13 / 21.04 dB.}
\label{fig:mcgetchin-ba}
\end{figure}

\subsection{Results and ablations}
\label{sec:ablations}
\input{figures/method-comparison/table}
\cref{tab:methods} gives the results of the baseline OrTeN configuration on each scene.
OrTeN attains sub-GSD height RMSE$_a$ in Jezero, Gale, and McGetchin scenes.
On the Olympus Mons scene, the RMSE$_a$ remains at 2.91 $\times$ GSD.
Its large relief (almost 3 km) and substantial cast shadows make reconstruction challenging. 
The lack of an independent, high-accuracy reference limits comparison with ASP, which performs similarly on this scene.

Depth-supervised EO-NeRF fails to reconstruct accurate DTMs in any of the scenes despite access to structural supervision.
It falls short of OrTeN both visually and in structural accuracy.
Noting that the default EO-NeRF configuration trains for much longer (300,000 iterations) and takes 10-20 hours to converge, this comparison demonstrates that OrTeN's approach achieves meaningful improvement in both efficiency and accuracy.

ASP recovers sound DTMs for all four scenes, but performs worse than OrTeN on all scenes except Olympus Mons.
as opposed to ASP, which typically processes images pairwise or in smaller subsets.
OrTeN also benefits from its semi-empirical appearance modeling which allows it to better deal with shadows and varying illumination conditions which may confound feature matching in purely photogrammetric approaches.

\input{figures/core-ablations/table}

Four ablation studies were conducted to evaluate the impact of the three key features in this work: bundle adjustment with jitter, normalized cross-correlation (NCC) for photometric consistency, and semi-empirical appearance modeling with shadow modeling.
The results of the ablation are summarized in \cref{tab:core-ablations}.

Removing BA significantly degrades the quality of the reconstructed DTMs across all scenes, highlighting the importance of joint optimization of camera parameters and scene geometry.
\cref{fig:mcgetchin-ba} shows baseline and no-BA renderings of the same acquisition for the McGetchin Crater scene.
The no-BA ablation shows clearly degraded visual reconstruction quality with a drop of more than 10 dB in reserved-pixel PSNR.
The DTM is similarly impacted with RMSE$_a$ increasing from 0.43 to 10.36 m and the quality of the reconstruction visibly degrading, especially in the fine detail around the central crater region.

\begin{figure}[!t]
\centering
\includegraphics[width=\columnwidth]{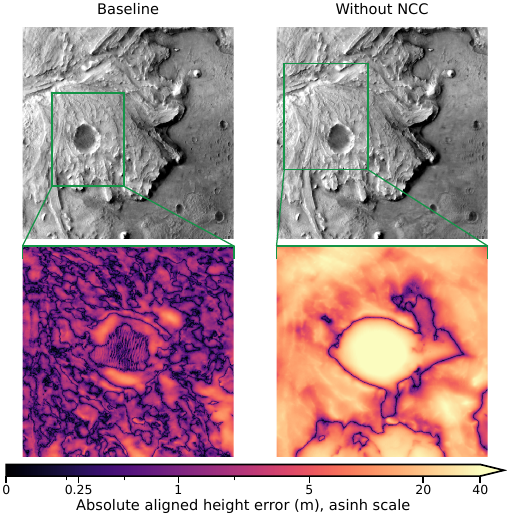}
\caption{\textbf{Jezero Crater with and without NCC.} 
Green boxes locate DTM error crops using each run's learned camera extrinsics and rigid alignment to the reference terrain. 
RMSE$_a$: 2.35 / 16.80 m. 
PSNR: 28.20 / 29.85 dB.}
\label{fig:jezero-ncc}
\end{figure}

Disabling NCC has an even more pronounced effect on the quality of the reconstructed DTMs.
\cref{fig:jezero-ncc} shows baseline and no-NCC renderings of the same acquisition for the Jezero Crater scene, and again illustrates the degradation of fine detail around the crater region.
In some of the scenes, the PSNR improves in the no-NCC ablation compared to the baseline.
Rendering loss alone can therefore be misleading when assessing terrain reconstruction, especially in low-disparity orbital scenes.

\begin{figure}[!t]
\centering
\includegraphics[width=\columnwidth]{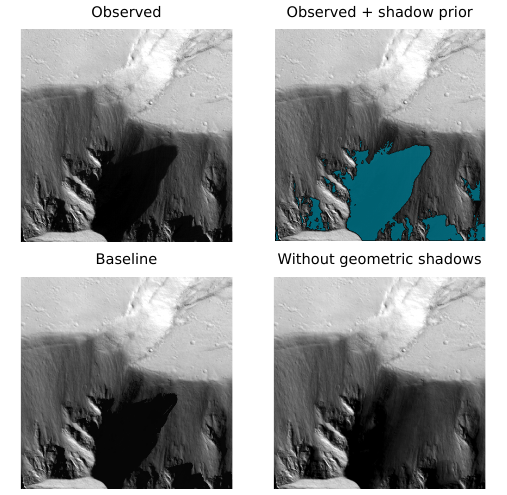}
\caption{\textbf{Olympus Mons shadow ablation.} 
Cyan marks shadows detected by the first-valley prior. 
Images share an intensity scale.}
\label{fig:olympus-shadows}
\end{figure}

Removal of geometric shadows produces a less pronounced effect on the quality of the reconstructed DTMs compared to the other ablations, but can be observed in reconstruction quality.
\cref{fig:olympus-shadows} shows baseline and no-shadow renderings of the same acquisition for the Olympus Mons scene.
The baseline is able to reconstruct the large shadow cast by the cliff face while the ablated scene cannot.

%% file: tables/datasets.tex
\begin{table}[!h]
\centering
\footnotesize
\color{black}
\hypersetup{allcolors=black}
\setlength{\tabcolsep}{2pt}
\begin{tabular*}{\columnwidth}{@{\extracolsep{\fill}}lrrlr@{}}
\toprule
Dataset & GSD & Band & Reference DTM & Acc. / prec. \\
 & (m/px) & (nm) & (post spacing) & (m) \\
\midrule
\shortstack[l]{Jezero$^a$\\Crater} & 6.41 & 500--700 & \shortstack[l]{Mars 2020\\(1 m)~\cite{fergason2020hirise}} & \shortstack[r]{$\sim$10--20 / \\$\sim$0.1--0.4~\cite{golombek2012selection}} \\
\shortstack[l]{Gale$^a$\\Crater} & 6.23 & 500--700 & \shortstack[l]{Gale mosaic\\(1 m)~\cite{calef_iii_msl_2016}} & \shortstack[r]{$\sim$10--20 / \\$\sim$0.1--0.4~\cite{golombek2012selection}} \\
\shortstack[l]{Olympus$^a$\\Mons} & 5.56 & 500--700 & \shortstack[l]{Audited ASP\\(11.12 m)} & --- / 11.68$^\ddagger$ \\
\shortstack[l]{McGetchin$^b$\\Crater} & 0.84 & 400--750 & \shortstack[l]{NAC\\NEW200CRAT3\\(3 m)~\cite{lroc2026new200}} & \shortstack[r]{1.51 / \\1.46~\cite{lroc2026new200}} \\
\bottomrule
\end{tabular*}
\caption{
\textbf{Dataset instrument and DTM reference overview.} 
Instruments are $^a$CTX~\cite{malin2007ctx} and $^b$LROC NAC~\cite{robinson2010lroc}. 
Acc. / prec. denotes vertical accuracy / precision.
GSD is estimated by intersecting adjacent pixel rays from initial cameras with the local ENU $xy$ plane.
$^\ddagger$For Olympus Mons, precision was estimated from the RMS elevation difference between two stereo DTMs generated from image pairs with no cameras in common during construction of the ASP reference used to evaluate OrTeN. 
}
\label{tab:datasets}
\end{table}

%% file: figures/method-comparison/table.tex
\begin{table}[t]
\centering
\footnotesize
\color{black}
\hypersetup{allcolors=black}
\setlength{\tabcolsep}{2pt}
\begin{tabular*}{\columnwidth}{@{\extracolsep{\fill}}lrr@{}}
\toprule
Scene / method & \shortstack{RMSE$_a$ / MAE$_a$ /\\slope RMSE$_a$} & PSNR / SSIM \\
 & (m / m / $^\circ$) & (dB / ---) \\
\midrule
\multicolumn{3}{@{}l}{Jezero Crater} \\
\quad \textbf{OrTeN} & \textbf{2.35} / \textbf{1.61} / \textbf{1.87} & \textbf{28.20} / \textbf{0.771} \\
\quad EO-NeRF + depth & 62.08 / 47.61 / 39.15 & 25.26 / 0.598 \\
\quad ASP & 4.90 / 3.56 / 4.81 & --- / --- \\
\midrule
\multicolumn{3}{@{}l}{Gale Crater} \\
\quad \textbf{OrTeN} & \textbf{2.82} / \textbf{2.00} / \textbf{2.54} & \textbf{31.37} / \textbf{0.840} \\
\quad EO-NeRF + depth & 156.37 / 125.21 / 27.04 & 29.64 / 0.719 \\
\quad ASP & 4.73 / 2.89 / 4.02 & --- / --- \\
\midrule
\multicolumn{3}{@{}l}{Olympus Mons} \\
\quad \textbf{OrTeN} & 16.18 / 8.57 / \textbf{3.43} & \textbf{29.71} / \textbf{0.904} \\
\quad EO-NeRF + depth & 25.76 / 20.14 / 9.01 & 16.87 / 0.708 \\
\quad ASP & \textbf{15.34} / \textbf{7.38} / 4.90 & --- / --- \\
\midrule
\multicolumn{3}{@{}l}{McGetchin Crater} \\
\quad \textbf{OrTeN} & \textbf{0.43} / \textbf{0.28} / \textbf{0.98} & \textbf{35.13} / \textbf{0.940} \\
\quad EO-NeRF + depth & 226.35 / 187.56 / 71.99 & 23.53 / 0.666 \\
\quad ASP & 0.56 / 0.38 / 1.14 & --- / --- \\
\bottomrule
\end{tabular*}
\caption{
\textbf{DTM reconstruction comparison.}
Best values per scene are bold.
Height errors are evaluated across each method's valid portion of the fixed evaluation footprint.
Slope uses local quadratic fits.
PSNR and SSIM use held-out pixels and a fixed intensity range per scene.
ASP produces no visual reconstruction.
}
\label{tab:methods}
\end{table}

%% file: figures/core-ablations/table.tex
\begin{table}[t]
\centering
\footnotesize
\color{black}
\hypersetup{allcolors=black}
\setlength{\tabcolsep}{2pt}
\renewcommand{\arraystretch}{1.12}
\begin{tabular*}{\columnwidth}{@{\extracolsep{\fill}}l*{4}{r}@{}}
\toprule
 & \multicolumn{4}{c}{RMSE$_a$ / PSNR} \\
Configuration & Jezero & Gale & Olympus & McGetchin \\
\midrule
Baseline & 2.35 / 28.20 & 2.82 / 31.37 & 16.18 / 29.71 & 0.43 / 35.13 \\
No BA & 4.26 / 26.40 & 9.25 / 28.39 & 30.01 / 24.31 & 10.36 / 21.04 \\
No jitter & 2.40 / 28.21 & 2.76 / 29.56 & 16.47 / 30.34 & 0.46 / 34.64 \\
No NCC & 16.80 / 29.85 & 42.22 / 29.09 & 67.40 / 22.06 & 0.52 / 34.63 \\
No shadows & 2.32 / 28.06 & 2.90 / 31.31 & 15.81 / 29.34 & 0.43 / 35.21 \\
\bottomrule
\end{tabular*}
\caption{\textbf{Core OrTeN ablations.} 
No BA also removes jitter. 
No shadows removes attached and cast visibility and the shadow prior.}
\label{tab:core-ablations}
\end{table}

%% file: sections/conclusion.tex
\section{Conclusion}
\label{sec:conclusion}
OrTeN jointly reconstructs terrain and refines line-scan cameras from calibrated orbital images without depth supervision or stereo initialization.
Across four Mars and Moon scenes, it achieves sub-GSD aligned terrain RMSE in three and improves on the tested ASP reconstruction in those same three scenes.
The ablations demonstrate the importance of camera refinement and surface-patch NCC for achieving high geometric accuracy.
They also show that an accurate image reconstruction can coexist with an incorrect surface.
Comparison with EO-NeRF with depth supervision also demonstrates that structural priors are not sufficient to achieve accurate terrain reconstruction efficiently.

However, a number of limitations remain.
Without structural priors such as ranging measurements or known control points, absolute geolocation is not possible.
Furthermore, it is possible that integrating structural priors into OrTeN could improve reconstruction accuracy and efficiency and should be tested.
While the height map representation is an effective structural prior for the datasets considered here, when detail becomes sufficient, this representation becomes more limiting.
For instance, height maps cannot represent overhangs or complex 3D structures like caves.
Future work should work on mitigating these shortcomings to allow ground level imagery to be integrated into the terrain reconstruction process.

%% file: main.bbl
\begin{thebibliography}{39}
\providecommand{\natexlab}[1]{#1}
\providecommand{\url}[1]{\texttt{#1}}
\expandafter\ifx\csname urlstyle\endcsname\relax
  \providecommand{\doi}[1]{doi: #1}\else
  \providecommand{\doi}{doi: \begingroup \urlstyle{rm}\Url}\fi

\bibitem[Adeline et~al.(2013)Adeline, Chen, Briottet, Pang, and
  Paparoditis]{adeline_shadow_2013}
K.~R.~M. Adeline, M. Chen, X. Briottet, S.~K. Pang, and N. Paparoditis.
\newblock Shadow detection in very high spatial resolution aerial images: A
  comparative study.
\newblock \emph{ISPRS Journal of Photogrammetry and Remote Sensing},
  80:\penalty0 21--38, 2013.

\bibitem[Aira et~al.(2025)Aira, Facciolo, and Ehret]{aira_gaussian_2025}
Luca~Savant Aira, Gabriele Facciolo, and Thibaud Ehret.
\newblock {Gaussian Splatting for Efficient Satellite Image Photogrammetry}.
\newblock In \emph{Proceedings of the IEEE/CVF Conference on Computer Vision
  and Pattern Recognition}, pages 5959--5969, 2025.

\bibitem[Alexandrov and Beyer(2018)]{alexandrov_multiview_2018}
Oleg Alexandrov and Ross~A. Beyer.
\newblock {Multiview Shape-From-Shading for Planetary Images}.
\newblock \emph{Earth and Space Science}, 5\penalty0 (10):\penalty0 652--666,
  2018.

\bibitem[{Ames Stereo Pipeline Developers}(2026)]{asp_jitter}
{Ames Stereo Pipeline Developers}.
\newblock {Ames Stereo Pipeline}: \texttt{jitter\_solve}.
\newblock Software documentation, version 3.7.0, 2026.
\newblock Accessed September 22, 2026.

\bibitem[{Arizona State University and Intuitive
  Machines}(2026)]{lroc2026new200}
{Arizona State University and Intuitive Machines}.
\newblock {NAC\_DTM\_NEW200CRAT3}: New 222 m crater stereo {DTM}.
\newblock LROC Planetary Data System product README, January 15, 2026.

\bibitem[Bell~{III} et~al.(2013)Bell~{III}, Malin, Caplinger, Fahle, Wolff,
  Cantor, James, Ghaemi, Posiolova, Ravine, Supulver, Calvin, Clancy, Edgett,
  Edwards, Haberle, Hale, Lee, Rice, and Williams]{bell_iii_calibration_2013}
J. Bell~{III}, M. Malin, M. Caplinger, J. Fahle, M. Wolff, B. Cantor, P. James,
  F. Ghaemi, L. Posiolova, Michael Ravine, K. Supulver, W. Calvin, R. Clancy,
  Kenneth Edgett, Laurence Edwards, R. Haberle, Amy Hale, S. Lee, Melissa Rice,
  and Rebecca Williams.
\newblock Calibration and performance of the mars reconnaissance orbiter
  context camera ({CTX}).
\newblock \emph{Mars}, 8:\penalty0 1--14, 2013.

\bibitem[Beyer et~al.(2018)Beyer, Alexandrov, and McMichael]{beyer_ames_2018}
Ross~A. Beyer, Oleg Alexandrov, and Scott McMichael.
\newblock {The Ames Stereo Pipeline: NASA's Open Source Software for Deriving
  and Processing Terrain Data}.
\newblock \emph{Earth and Space Science}, 5\penalty0 (9):\penalty0 537--548,
  2018.

\bibitem[Biberstein et~al.(2025)Biberstein, Cavalheiro, Han, and
  Karaman]{biberstein2025ntm}
Josef~X. Biberstein, Guilherme Cavalheiro, Juyeop Han, and Sertac Karaman.
\newblock Construction of digital terrain maps from multi-view satellite
  imagery using neural volume rendering.
\newblock \emph{arXiv preprint arXiv:2508.01386}, 2025.

\bibitem[Bournez et~al.(2025)Bournez, Aira, Ehret, and
  Facciolo]{bournez_eogs_2025}
Pierrick Bournez, Luca~Savant Aira, Thibaud Ehret, and Gabriele Facciolo.
\newblock {EOGS++: Earth Observation Gaussian Splatting with Internal Camera
  Refinement and Direct Panchromatic Rendering}, 2025.

\bibitem[Calef~III and Parker(2016)]{calef_iii_msl_2016}
F.~J. Calef~III and T. Parker.
\newblock {MSL Gale Merged Orthophoto Mosaic}, 2016.

\bibitem[Cheng et~al.(2021)Cheng, Ansar, and Johnson]{cheng2021reference}
Yang Cheng, Adnan Ansar, and Andrew Johnson.
\newblock Making an onboard reference map from {MRO/CTX} imagery for {Mars
  2020} lander vision system.
\newblock \emph{Earth and Space Science}, 8\penalty0 (8):\penalty0
  e2020EA001560, 2021.

\bibitem[de~Almeida et~al.(2026)de~Almeida, Brydon, Persaud, Williamson, and
  Henderson]{almeida_neural_2026}
Melonie de Almeida, George Brydon, Divya~M. Persaud, John~H. Williamson, and
  Paul Henderson.
\newblock Neural 3d reconstruction of planetary surfaces from descent-phase
  wide-angle imagery.
\newblock In \emph{Proceedings of the IEEE/CVF Conference on Computer Vision
  and Pattern Recognition Workshops}, pages 10174--10182, 2026.

\bibitem[de~Franchis et~al.(2014{\natexlab{a}})de~Franchis, Meinhardt-Llopis,
  Michel, Morel, and Facciolo]{de_franchis_automatic_2014}
C. de Franchis, E. Meinhardt-Llopis, J. Michel, J.-M. Morel, and G. Facciolo.
\newblock {An automatic and modular stereo pipeline for pushbroom images}.
\newblock \emph{ISPRS Annals of the Photogrammetry, Remote Sensing and Spatial
  Information Sciences}, II-3:\penalty0 49--56, 2014{\natexlab{a}}.

\bibitem[de~Franchis et~al.(2014{\natexlab{b}})de~Franchis, Meinhardt-Llopis,
  Michel, Morel, and Facciolo]{defranchis_on_2014}
C. de Franchis, E. Meinhardt-Llopis, J. Michel, J.-M. Morel, and G. Facciolo.
\newblock {On stereo-rectification of pushbroom images}.
\newblock In \emph{2014 IEEE International Conference on Image Processing
  (ICIP)}, pages 5447--5451, 2014{\natexlab{b}}.

\bibitem[Derksen and Izzo(2021)]{derksen_shadow_2021}
Dawa Derksen and Dario Izzo.
\newblock {Shadow Neural Radiance Fields for Multi-view Satellite
  Photogrammetry}, 2021.

\bibitem[Fergason et~al.(2020)Fergason, Galuszka, Hare, Mayer, and
  Redding]{fergason2020hirise}
Robin~L. Fergason, Donna~M. Galuszka, Trent~M. Hare, David~P. Mayer, and
  Bonnie~L. Redding.
\newblock {Mars 2020} terrain relative navigation {HiRISE DTM} mosaic.
\newblock U.S. Geological Survey data release, 2020.

\bibitem[Gableman and Kak(2024)]{gableman_incorporating_2024}
Michael Gableman and Avinash Kak.
\newblock {Incorporating Season and Solar Specificity Into Renderings Made by a
  NeRF Architecture Using Satellite Images}.
\newblock \emph{IEEE Transactions on Pattern Analysis and Machine
  Intelligence}, 46\penalty0 (6):\penalty0 4348--4365, 2024.

\bibitem[Gao et~al.(2012)Gao, Burt, and Zhu]{gao2012slope}
Jing Gao, James~E. Burt, and A-Xing Zhu.
\newblock Neighborhood size and spatial scale in raster-based slope
  calculations.
\newblock \emph{International Journal of Geographical Information Science},
  2012.

\bibitem[Geng et~al.(2020)Geng, Xu, Xing, and Lan]{geng_generic_2020}
Xun Geng, Qing Xu, Shuai Xing, and Chaozhen Lan.
\newblock {A Generic Pushbroom Sensor Model for Planetary Photogrammetry}.
\newblock \emph{Earth and Space Science}, 7\penalty0 (5):\penalty0
  e2019EA001014, 2020.

\bibitem[Golombek et~al.(2012)Golombek, Grant, Kipp, Vasavada, Kirk, Fergason,
  et~al.]{golombek2012selection}
Matthew Golombek, John~A. Grant, D. Kipp, A. Vasavada, R. Kirk, R. Fergason,
  et~al.
\newblock Selection of the mars science laboratory landing site.
\newblock \emph{Space Science Reviews}, 170:\penalty0 641--737, 2012.

\bibitem[Jung et~al.(2020)Jung, Lee, and Bang]{jung_digital_2020}
Youeyun Jung, Seongheon Lee, and Hyochoong Bang.
\newblock {Digital Terrain Map Based Safe Landing Site Selection for Planetary
  Landing}.
\newblock \emph{IEEE Transactions on Aerospace and Electronic Systems},
  56\penalty0 (1):\penalty0 368--380, 2020.

\bibitem[{Kelvin Rodriguez} and {Astrogeology Science
  Center}(2024)]{kelvin_rodriguez_integrated_2024}
{Kelvin Rodriguez} and {Astrogeology Science Center}.
\newblock Integrated software for imagers and spectrometers ({ISIS}) 8.3.0,
  2024.

\bibitem[Lee et~al.(2026)Lee, Yim, Park, Kim, Kim, Jeong, Sim, and
  Lee]{lee_lnem_nodate}
Suwan Lee, Jo~Ryeong Yim, Kibaek Park, Dong-Gyu Kim, Eunhyeuk Kim, Minsup
  Jeong, Chae~Kyung Sim, and Seokju Lee.
\newblock {LNEM}: Lunar neural elevation model.
\newblock In \emph{Proceedings of the IEEE/CVF Conference on Computer Vision
  and Pattern Recognition}, pages 6508--6517, 2026.

\bibitem[Lin et~al.(2021)Lin, Ma, Torralba, and Lucey]{lin2021barf}
Chen-Hsuan Lin, Wei-Chiu Ma, Antonio Torralba, and Simon Lucey.
\newblock {BARF}: Bundle-adjusting neural radiance fields.
\newblock In \emph{Proceedings of the IEEE/CVF International Conference on
  Computer Vision}, 2021.

\bibitem[Malin et~al.(2007)Malin, Bell, Cantor, Caplinger, Calvin, Clancy,
  Edgett, Edwards, Haberle, James, Lee, Ravine, Thomas, and
  Wolff]{malin2007ctx}
Michael~C. Malin, James~F. Bell, Bruce~A. Cantor, Michael~A. Caplinger,
  Wendy~M. Calvin, R.~Todd Clancy, Kenneth~S. Edgett, Laurence Edwards,
  Robert~M. Haberle, Philip~B. James, Steven~W. Lee, Michael~A. Ravine,
  Peter~C. Thomas, and Michael~J. Wolff.
\newblock Context camera investigation on board the mars reconnaissance
  orbiter.
\newblock \emph{Journal of Geophysical Research: Planets}, 112:\penalty0
  E05S04, 2007.

\bibitem[Mari et~al.(2022)Mari, Facciolo, and Ehret]{mari_sat-nerf_2022}
Roger Mari, Gabriele Facciolo, and Thibaud Ehret.
\newblock {Sat-NeRF: Learning Multi-View Satellite Photogrammetry With
  Transient Objects and Shadow Modeling Using RPC Cameras}.
\newblock In \emph{2022 IEEE/CVF Conference on Computer Vision and Pattern
  Recognition Workshops (CVPRW)}, pages 1310--1320. IEEE, 2022.

\bibitem[Mar{\'{i}} et~al.(2023)Mar{\'{i}}, Facciolo, and
  Ehret]{mari_multi-date_2023}
Roger Mar{\'{i}}, Gabriele Facciolo, and Thibaud Ehret.
\newblock {Multi-Date Earth Observation NeRF: The Detail Is in the Shadows}.
\newblock In \emph{2023 IEEE/CVF Conference on Computer Vision and Pattern
  Recognition Workshops (CVPRW)}, pages 2035--2045. IEEE, 2023.

\bibitem[McEwen(1991)]{mcewen_photometric_1991}
Alfred~S. McEwen.
\newblock Photometric functions for photoclinometry and other applications.
\newblock \emph{Icarus}, 92\penalty0 (2):\penalty0 298--311, 1991.

\bibitem[McEwen et~al.(2007)McEwen, Eliason, Bergstrom, Bridges, Hansen,
  Delamere, Grant, Gulick, Herkenhoff, Keszthelyi, Kirk, Mellon, Squyres,
  Thomas, and Weitz]{mcewen2007hirise}
Alfred~S. McEwen, Eric~M. Eliason, James~W. Bergstrom, Nathan~T. Bridges,
  Candice~J. Hansen, W.~Alan Delamere, John~A. Grant, Virginia~C. Gulick,
  Kenneth~E. Herkenhoff, Laszlo Keszthelyi, Randolph~L. Kirk, Michael~T.
  Mellon, Steven~W. Squyres, Nicolas Thomas, and Catherine~M. Weitz.
\newblock {Mars Reconnaissance Orbiter}'s {High Resolution Imaging Science
  Experiment (HiRISE)}.
\newblock \emph{Journal of Geophysical Research: Planets}, 112:\penalty0
  E05S02, 2007.

\bibitem[M{\"{u}}ller et~al.(2022)M{\"{u}}ller, Evans, Schied, and
  Keller]{muller_instant_2022}
Thomas M{\"{u}}ller, Alex Evans, Christoph Schied, and Alexander Keller.
\newblock {Instant Neural Graphics Primitives with a Multiresolution Hash
  Encoding}.
\newblock \emph{ACM Transactions on Graphics}, 41\penalty0 (4):\penalty0 1--15,
  2022.

\bibitem[Qu and Deng(2023)]{qu_sat-mesh_2023}
Yingjie Qu and Fei Deng.
\newblock {Sat-Mesh: Learning Neural Implicit Surfaces for Multi-View Satellite
  Reconstruction}.
\newblock \emph{Remote Sensing}, 15\penalty0 (17):\penalty0 4297, 2023.

\bibitem[Robinson et~al.(2010)Robinson, Brylow, Tschimmel, Humm, Lawrence,
  Thomas, Denevi, Bowman-Cisneros, Zerr, Ravine, Caplinger, Ghaemi, Schaffner,
  Malin, Mahanti, Bartels, Anderson, Tran, Eliason, McEwen, Turtle, Jolliff,
  and Hiesinger]{robinson2010lroc}
M.~S. Robinson, S.~M. Brylow, M. Tschimmel, D. Humm, S.~J. Lawrence, P.~C.
  Thomas, B.~W. Denevi, E. Bowman-Cisneros, J. Zerr, M.~A. Ravine, M.~A.
  Caplinger, F.~T. Ghaemi, J.~A. Schaffner, M.~C. Malin, P. Mahanti, A.
  Bartels, J. Anderson, T.~N. Tran, E.~M. Eliason, A.~S. McEwen, E. Turtle,
  B.~L. Jolliff, and H. Hiesinger.
\newblock {Lunar Reconnaissance Orbiter Camera (LROC)} instrument overview.
\newblock \emph{Space Science Reviews}, 150:\penalty0 81--124, 2010.

\bibitem[Sch{\"{o}}nberger and Frahm(2016)]{schonberger_structure--motion_2016}
Johannes~Lutz Sch{\"{o}}nberger and Jan-Michael Frahm.
\newblock {Structure-from-Motion Revisited}.
\newblock In \emph{Conference on Computer Vision and Pattern Recognition
  (CVPR)}, 2016.

\bibitem[Sch{\"{o}}nberger et~al.(2016)Sch{\"{o}}nberger, Zheng, Pollefeys, and
  Frahm]{schonberger_pixelwise_2016}
Johannes~Lutz Sch{\"{o}}nberger, Enliang Zheng, Marc Pollefeys, and Jan-Michael
  Frahm.
\newblock {Pixelwise View Selection for Unstructured Multi-View Stereo}.
\newblock In \emph{European Conference on Computer Vision (ECCV)}, 2016.

\bibitem[Tancik et~al.(2023)Tancik, Weber, Ng, Li, Yi, Kerr, Wang,
  Kristoffersen, Austin, Salahi, Ahuja, McAllister, and
  Kanazawa]{tancik_nerfstudio_2023}
Matthew Tancik, Ethan Weber, Evonne Ng, Ruilong Li, Brent Yi, Justin Kerr,
  Terrance Wang, Alexander Kristoffersen, Jake Austin, Kamyar Salahi, Abhik
  Ahuja, David McAllister, and Angjoo Kanazawa.
\newblock {Nerfstudio: A Modular Framework for Neural Radiance Field
  Development}.
\newblock In \emph{Special Interest Group on Computer Graphics and Interactive
  Techniques Conference Conference Proceedings}, pages 1--12, 2023.

\bibitem[Wagner et~al.(2026)Wagner, Bullinger, Arens, and
  Stiefelhagen]{wagner_satgeo_2026}
Valentin Wagner, Sebastian Bullinger, Michael Arens, and Rainer Stiefelhagen.
\newblock {SatGeo-NeRF}: Geometrically regularized {NeRF} for satellite
  imagery.
\newblock \emph{ISPRS Annals of the Photogrammetry, Remote Sensing and Spatial
  Information Sciences}, XI-2-2026:\penalty0 367--374, 2026.

\bibitem[Wang et~al.(2021)Wang, Liu, Liu, Theobalt, Komura, and
  Wang]{wang_neus_2023}
Peng Wang, Lingjie Liu, Yuan Liu, Christian Theobalt, Taku Komura, and Wenping
  Wang.
\newblock {NeuS: Learning Neural Implicit Surfaces by Volume Rendering for
  Multi-view Reconstruction}.
\newblock In \emph{Advances in Neural Information Processing Systems}, pages
  27171--27183. Curran Associates, Inc., 2021.

\bibitem[Zhang and Rupnik(2023)]{zhang_sparsesat-nerf_2023}
Lulin Zhang and Ewelina Rupnik.
\newblock {SparseSat-NeRF: Dense Depth Supervised Neural Radiance Fields for
  Sparse Satellite Images}, 2023.

\bibitem[Zhang et~al.(2025)Zhang, Rupnik, Nguyen, Jacquemoud, and
  Klinger]{zhang_brdf_2025}
Lulin Zhang, Ewelina Rupnik, Tri~Dung Nguyen, St{\'{e}}phane Jacquemoud, and
  Yann Klinger.
\newblock {BRDF-NeRF: Neural radiance fields with optical satellite images and
  BRDF modelling}.
\newblock \emph{International Journal of Applied Earth Observation and
  Geoinformation}, 143:\penalty0 104747, 2025.

\end{thebibliography}
